\documentclass[letterpaper, 10 pt, conference]{ieeeconf}  

\IEEEoverridecommandlockouts                              

\usepackage{graphics} 
\usepackage{graphicx}
\usepackage{subcaption}
\usepackage{amsmath} 
\usepackage{amssymb}  
\usepackage{float}
\title{\LARGE \bf
Extending Policy from One-Shot Learning through Coaching}

\author{Mythra V. Balakuntala$^{1}$, Vishnunandan L. N. Venkatesh$^{1}$, Jyothsna Padmakumar Bindu$^{1}$,\\ Richard M. Voyles$^{1}$, Juan Wachs$^{2}$
\thanks{*This work was supported by the NSF Center on RObots and SEnsors for HUman well-Being (RoSe-HuB) under grant CNS-1439717 and by the Office of the Assistant Secretary of Defense for Health Affairs under Award No. W81XWH-18-1-0769.}
\thanks{$^{1}$School of Engineering Technology, $^{2}$School of Industrial Engineering, 
        Purdue University, IN 47907, USA
        {\tt\small mbalakun@purdue.edu, lvenkate@purdue.edu, jpadmaku@purdue.edu, rvoyles@purdue.edu, jpwachs@purdue.edu }}%
        }
\begin{document}

\maketitle
\thispagestyle{empty}
\pagestyle{empty}
\begin{abstract}
Humans generally teach their fellow collaborators to perform tasks through a small number of  demonstrations. The learnt task is corrected or extended to meet specific task goals by means of coaching. Adopting a similar framework for teaching robots through demonstrations and coaching makes teaching tasks highly intuitive. Unlike traditional Learning from Demonstration (LfD) approaches which require multiple demonstrations, we present a one-shot learning from demonstration approach to learn tasks. The learnt task is corrected and generalized using two layers of evaluation/modification. First, the robot self-evaluates its performance and corrects the performance to be closer to the demonstrated task. Then, coaching is used as a means to extend the policy learnt to be adaptable to varying task goals. Both the self-evaluation and coaching are implemented using reinforcement learning (RL) methods. Coaching is achieved through human feedback on desired goal and action modification to generalize to specified task goals. The proposed approach is evaluated with a scooping task, by presenting a single demonstration. The self-evaluation framework aims to reduce the resistance to scooping in the media. To reduce the search space for RL, we bootstrap the search using least resistance path obtained using resistive force theory. Coaching is used to generalize the learnt task policy to transfer the desired quantity of material. Thus, the proposed method provides a framework for learning tasks from one demonstration and generalizing it using human feedback through coaching. 
\end{abstract}
\section{INTRODUCTION}

  Adaptability to new uncertain settings is a primary requirement for social collaborative robots in domestic environments like homes, hospitals, restaurant or offices. The complexity and unpredictability of these environments require robots to be able to learn, and adapt on the fly in contrast to performing pre-programmed tasks. This can be achieved through leveraging knowledge of tasks in the form of demonstrations from task experts (humans) \cite{billard2016learning} who can work in such environments with ease. These approaches are broadly referred to as learning from demonstration (LfD) or programming by demonstration (PbD). LfD and PbD are very intuitive to humans and require no knowledge about robot configuration or programming, hence can be done with little to no training. 
  
  Most tasks in domestic settings requires the robot to interact with the environment, making them contact-intensive or force-based tasks rather than a pure kinematic task. Earlier LfD approaches based on play-back, or kinematic approaches fail to learn such force-based tasks. One approach to obtain forces required for the task from demonstration is by choosing interfaces which capture the force information. Hence the interface used for demonstrating the task plays a crucial role in LfD approaches, it determines the richness of data available to learn the task. Interfaces like sensorized gloves \cite{voyles1999adaptation}, or kinesthetic teaching (hand-held guiding) \cite{hersch2008dynamical} enables the robot to acquire sufficient information to learn force signatures of tasks during the demonstration. But are not intuitive for the human demonstrator. Hence, in this work, a vision interface (Kinect RGBD camera) is used as the means to acquire the demonstrations in a natural way.
\begin{figure}[t]
    \centering
    \includegraphics[width=0.46\textwidth]{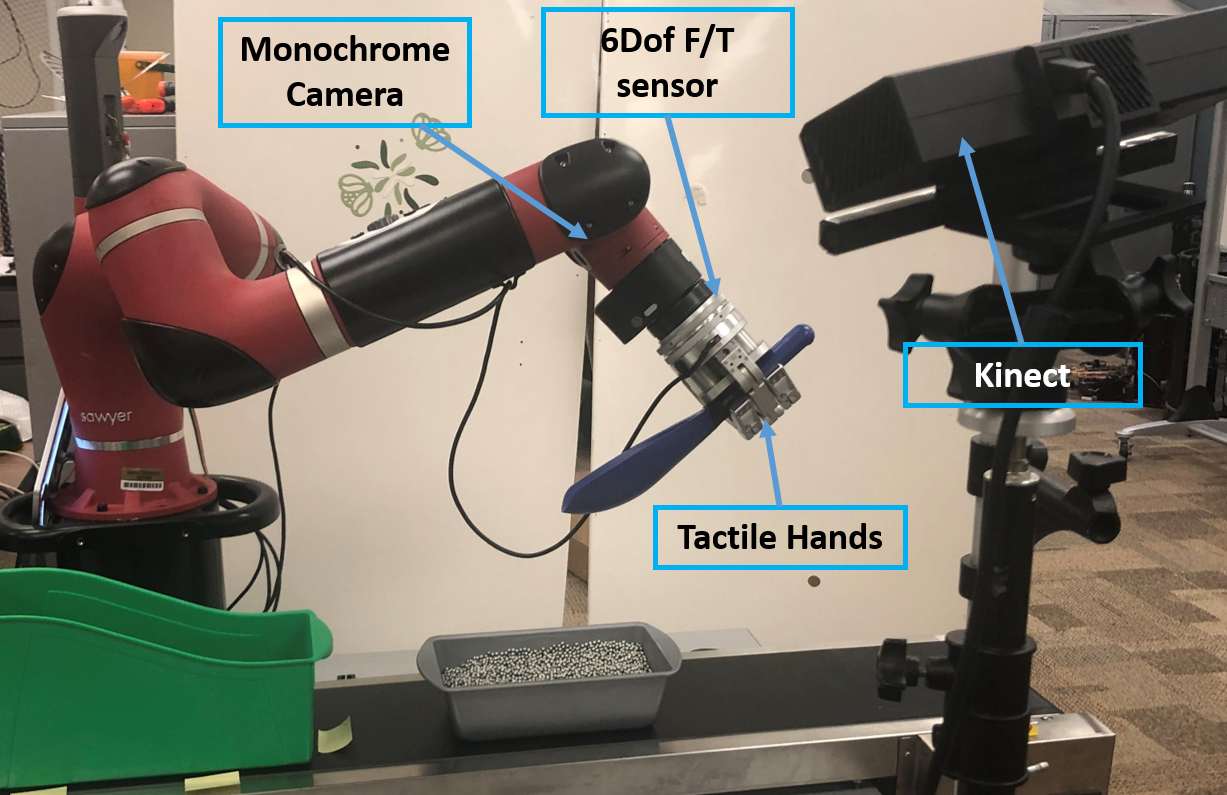}
    \caption{Super Baxter robot testbed with multimodal sensors}
    \label{fig:rob}
\end{figure}

  The data obtained from the vision interface are encoded into policies through supervised learning or reinforcement learning (RL) methods where demonstrations are used to bootstrap the RL. Supervised learning methods require large number of demonstrations to obtain a good policy which can generalize. But, it is not user friendly to record multiple demonstrations for each new task, meaning the robot should learn from few demonstrations, ideally one. These approaches are called one-shot learning methods. This work builds on the one-shot policy learning approach for contact-intensive tasks presented in \cite{balakuntala2019self}, where a task policy is learnt as a composition of a priori skills. Using a visual interface and one demonstration to learn from enables a natural and intuitive interface for task experts to program the robot through demonstration. But, inferring force information from visual demonstrations is not feasible and if the provided demonstration is suboptimal, the learn policy will not behave as desired. To overcome these limitations, a two-level stratified evaluation-correction framework is proposed. The fist layer is self-evaluation, where the robot evaluates the task performance against observed demonstration and adapts the skill parameters like force signatures, through a RL method \cite{balakuntala2019self}. Additional domain information like force models for fluids and granular media is provided to estimate baseline force signatures which are used to reduce or bias the RL action space.

  If the demonstration is suboptimal or is noisy, the learnt policy after self-evaluation is still inaccurate, hence requiring another layer of policy correction. This is done through the process of coaching. The user provides feedback for the task, which signifies a desired correction. Coaching can also be a means to generalize the task to a different goal quantity. For example, the task can be teaching a person to throw a basketball into the hoop. After teaching the basic technique for throwing a ball, the coach will then provide additional modifications to angle of throw and launch velocity to adapt the throwing to distance from the basket. Hence, coaching can be a means of correcting policy or generalizing the policy to a wider setting. Human-in-the-loop RL approaches use feedback directly as the reward, but this has been shown to be ineffective  \cite{macglashan2017interactive}. Instead, this work presents an approach where the feedback is used to specify task goal which is used to create the reward function, thus modifying the base policy learnt from demonstration. Hence, there is two levels of policy adaptation, the self evaluation step which tries to achieve policy to match observed demonstration, and coaching step which aims to generalize or correct the errors in demonstration. The proposed framework is evaluated with scooping granular media as the task, where self-evaluation layer attempts to achieve scooping with least effort (along minimum resistance path) and coaching is used to adapt the task to unscoop specific quantity.

\section{RELATED WORK}
Different LfD architectures have been explored that employ various machine learning techniques, such as Gaussian Mixture Models, Hidden Markov Models, Neural Networks, Dynamic Motion Primitives, and Reinforcement Learning \cite{billard2016learning}, \cite{argall2011teacher}, \cite{stulp2012reinforcement}. Methods involving inverse optimal control or Inverse Reinforcement Learning \cite{billard2016learning}, \cite{abbeel2004apprenticeship}  are also used in cases where RL policy rewards are inferred from the demonstrations. Learning from Demonstration paradigms have been used to learn various kinematic tasks which mostly involve pick and place operations. Contact intensive tasks \cite{elliott2017learning} cannot be easily inferred through noisy visual demonstrations, which adds additional layer of complexity. Solutions to these limitations require providing prior knowledge of environment, i.e. states and skills \cite{levine2015learning}.  A generalized approach is proposed in \cite{balakuntala2019self} to tackling these limitations. A gross policy is learnt from demonstration based on physical interactions among agent and objects, and the corresponding forces are self learnt by the robot using reinforcement learning.

LfD paradigms today have transitioned to using only few to one demonstration to infer the task. In these cases, methods like inverse reinforcement learning would not help as these methods require large amounts of data. A one-shot gesture recognition method is presented in \cite{cabrera2017one}, where multiple instances can be created from one example video by using Gaussian mixture regression, around inflection points in the trajectory. One shot learning methods \cite{finn2017one} have also been used to infer the task directly from a single demonstration using policies learnt for related tasks. Using only one demonstration can add complexities when it comes to inference. The robot would solely try to learn the task from one demonstration and this further increases the ambiguity in task inferences that can be made to achieve the optimal policy. This weakness can be overcome by using human in loop coaching methods or human in loop evaluation. 

Early works in coaching include various forms of human feedback based reinforcement learning methods \cite{thomaz2005real}. In these RL methods a human trainer can provide a reward to the robot agent during the agent's policy execution. Most of the applications involving human in loop RL involve the humans assigning rewards to the agent. Other actor critic \cite{macglashan2017interactive} methods have also been used to perform coaching using policy dependent feedback methods. Kinesthetic coaching methods \cite{calinon2007teacher}, \cite{calinon2007incremental} involve having the robot learn from demonstrations and then the human teacher refines the robot's learnt policy by manually making adjustments on the robot kinesthetically during its policy execution. More explorative work uses multiple forms of human coaching to determine which method among visual feedback, positional feedback and force feedback works best among subjects using the coaching interface \cite{gams2016line}. The results from our self-skill-evaluation experiments indicate that introducing coaching in our LfD approach \cite{balakuntala2019self} shall improve task performance. Human-in-the-loop evaluation as in \cite{mollard2015robot} can help us obtain more accurate skill parameters as well as better transition between skills. In  the context of coaching \cite{gams2016line} tells us how coaching vs learning plays a crucial role. Coaching would perform best to only tune the already learnt policy but is not a replacement to learning the complete policy itself. The methods proposed in this paper deals with taking human feedback for coaching which would help the robot achieve specific task goals using the inferred skill from a single demonstration. 

\section{Task Learning from demonstration}
\begin{figure*}[t]
    \centering
    \includegraphics[width=0.6\textwidth]{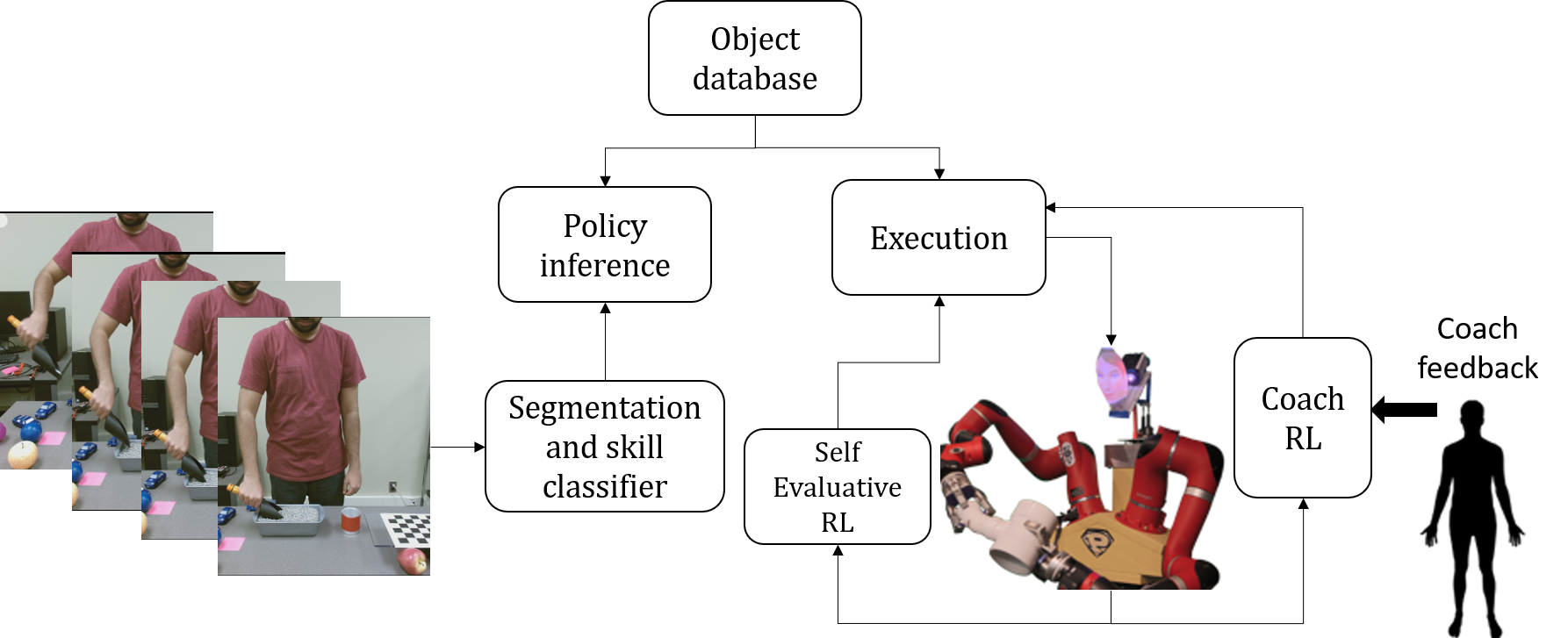}
    \caption{The proposed one-shot learning with coaching framework for our Super Baxter robot \cite{soratana2018SuperBaxter}}
    \label{fig:arch}
\end{figure*}

In this section, we describe the learning from demonstration paradigm proposed in this paper. The learning begins with one visual demonstration of the task. A task policy is learnt from this single demonstration as a sequence of sensorimotor primitives or skills that the robot knows to perform. Then the self-evaluation layer performs skill tuning to achieve observed policy. Finally a coaching layer is explained, which incorporates user feedback to extend the learnt policy. The entire framework of the proposed approach is shown in Fig. \ref{fig:arch}.
\subsection{Inference from vision}
The demonstration obtained is in the form of a single RGB-D video recorded using Microsoft KINECT. The robot has a pre-learnt database of objects, containing features and possible states for different objects. The features include shape, mass, stiffness, shape signatures and states like peeled or unpeeled for vegetables, mass, grasp orientation and  filled/empty status of containers. The inference framework used in this work is the same as one presented in \cite{balakuntala2019self}, which is extended to incorporate coaching in the framework. This one-shot task learning is inspired by other one-shot methods which use characteristic key points \cite{wan2016explore}, \cite{cabrera2017one}. The inference framework uses physical interaction keypoints (PIKs) to segment the demonstration into skill (sensorimotor primitives). PIKs signify points on the hand trajectory where there is change in contact condition between agent and object or object in hand and another object. Two binary features $\phi_i$ and $\psi_i$ are computed for each segment which indicate contact (1) or lack thereof (0), between hand-nearest object and object in hand-nearest object respectively. The PIKs indicate the change in skills and are used for temporal segmentation of the entire demonstration into multiple segments $\Theta_i$, where each PIK represents the point of segmentation. For each of the segments, relative motion trajectories $X_i$ between hand and the object it is interacting with are extracted. If the hand is not interacting with any object i.e. $\phi_i = 0$ then the relative motion trajectories are computed with the object it interacts with in segment $\Theta_{i+1}$. Then each segment is represented as $$\Theta_i = (\psi_i, X_i, u(\dot{X}_i), u(\dot{Y}_i), \phi_i)$$ where $Y_i$ is the absolute velocity of the hand, and
\begin{eqnarray}
u(\dot{X}_i) = \begin{cases} 1 & \text{if}\quad \dot{X}_i \geq 0\\
0 & \text{otherwise}\end{cases}
\end{eqnarray}
Let the skill class label of segment $\Theta_i$ be $C_i$. Then the set of $\phi_i$, $\psi_i$, $u(\dot{X}_i)$, the class label of previous class $C_{i-1}$, and object ID values is used to classify each segment $\Theta_i$ into a priori skill using decision tree classifier.

The learnt policy is obtained in two steps - firstly, the sequence of the inferred skill classes is obtained as above. Then any necessary skills required for transition from skill $C_i$ to $C_{i+1}$ are added into the sequence. The final learnt policy is the sequence of the a priori skills,
\begin{equation*}\Pi = \{(C_1, s*_1, \Theta_1), (C_2, s*_2, \Theta_2), \dots (C_m, s*_m, \Theta_m)\}\end{equation*}
 where $s*_i$ is the state of object agent is interacting with after skill $C_i$ in the demonstration. This state is the reference goal state for the skill execution. $m$ is the number of skills and $\Theta_i$ the $i\textsuperscript{th}$ demonstration segment. Each skill $C_i$ is associated with an execution sensorimotor control model to perform the skill. This is described in the following section.

\subsection{A priori skills} \label{skills}
\begin{figure}[t]
    \centering
    \begin{subfigure}[t]{0.1\textwidth}
        \centering
        \includegraphics[width=\textwidth]{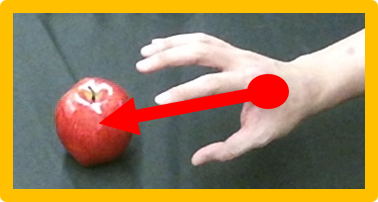}
        \caption{Approach}
    \end{subfigure}
    \begin{subfigure}[t]{0.1\textwidth}
        \centering
        \includegraphics[width=\textwidth]{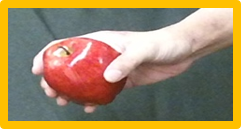}
        \caption{Grasp}
    \end{subfigure}
    \begin{subfigure}[t]{0.1\textwidth}
        \centering
        \includegraphics[width=\textwidth]{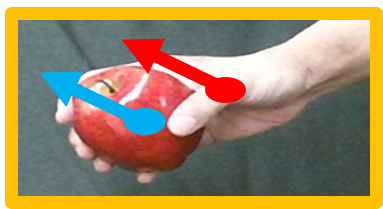}
        \caption{Transport}
    \end{subfigure}
    \begin{subfigure}[t]{0.1\textwidth}
        \centering
        \includegraphics[width=\textwidth]{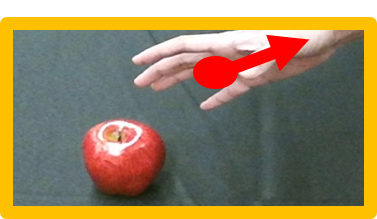}
        \caption{Retract}
    \end{subfigure}
    \begin{subfigure}[t]{0.1\textwidth}
        \centering
        \includegraphics[width=\textwidth]{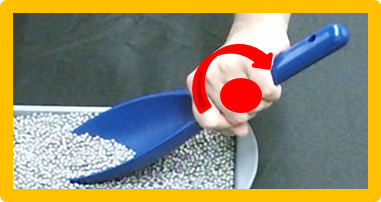}
        \caption{Scoop}
    \end{subfigure}
    \begin{subfigure}[t]{0.1\textwidth}
        \centering
        \includegraphics[width=\textwidth]{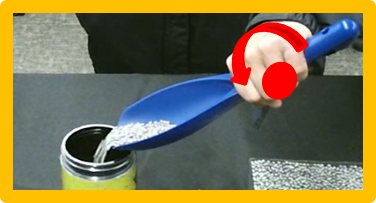}
        \caption{Unscoop}
    \end{subfigure}
    \begin{subfigure}[t]{0.1\textwidth}
        \centering
        \includegraphics[width=\textwidth]{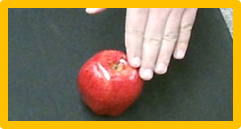}
        \caption{Guarded Move}
    \end{subfigure}
    \begin{subfigure}[t]{0.1\textwidth}
        \centering
        \includegraphics[width=\textwidth]{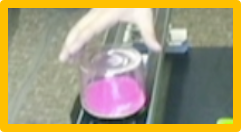}
        \caption{Visual servoing}
    \end{subfigure}
    \begin{subfigure}[t]{0.1\textwidth}
        \centering
        \includegraphics[width=0.8\textwidth]{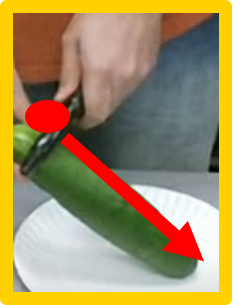}
        \caption{Move with contact}
    \end{subfigure}
    \begin{subfigure}[t]{0.1\textwidth}
        \centering
        \includegraphics[width=0.8\textwidth]{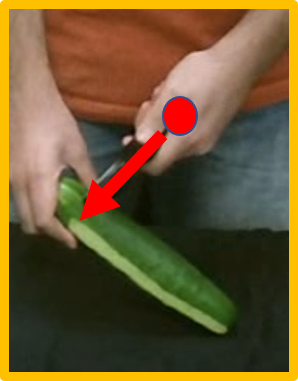}
        \caption{Move to contact}
    \end{subfigure}
    \caption{List of a priori skills.}
    \label{fig:skills}
\end{figure}

A priori skills are atomic sensorimotor control actions which the robot can perform. Fig \ref{fig:skills} shows some of these skills. Each skill is defined  as a control action with a particular sensor feedback and goal condition. The database of skills is segmented into two types - force-based and positional. Force-based skills have an impedance control policy for achieving desired force trajectories and positional skills have a positional control policy as shown in \ref{eqn:ski}.
Let the state of the system be denoted by $s$, the pose at time $t$ by $x_t$, the goal state by $s*$, and the desired pose by $x^d$. Then, the skill control policy for kinematic skills can be defined as follows,
\begin{equation}
    x_{t+1} = x_t + k_1(f(x^d, s*) + k_2 \dot{f}(x^d,s*)) \label{eqn:ski}
\end{equation},
where $k_1$ and $k_2$ are gain parameters, $f$ is the total sensor feedback error function which depends on the desired pose $x_d$ and goal state. 
We use an impedance control policy for force-based skills like \textit{move with contact} as follows, 
\begin{equation}
    \tau = J(\theta_t)^T F_d + K_1(f(x^d, s*) + K_2 \dot{f}(x^d,s*)) \label{eqn:fski}
\end{equation}
where $\tau$ is the 7-dof joint torque vector, $J(\theta_t)$ is the Jacobian at joint configuration $\theta_t$ and $F_d$ is the desired pose at the end-effector. The feedback error function is a combination of feedback from the different sensing modalities based on the skill being used like tactile forces for \textit{grasp} skill and vision for \textit{visual servoing}.

\subsection{Inferring force signatures} For tasks that involve manipulation in any other media than air, like stirring, scooping granular media, digging, humans generally adapt a goal driven least resistance or least effort path. Instead of the skill reinforcement learner searching the entire space for least resistance path, we bias the action space based on expected theoretical least effort path estimated using force models. Essentially, the action space is constrained to be bounded orientation changes around the least resistance path, to account for assumptions in the force model. The least effort path in media is computed using a variational approach, where the path is the functional to be estimated. Force in a granular or fluid media can be modelled using resistive force theory \cite{zhang2014effectiveness} as shown in the equation below.
\begin{equation}
    F = \int \left(dF_\perp + d F_{||} \right) \label{eqn:rft}
\end{equation}
where $dF_\perp$ is the resistive force normal to the direction of motion and $dF_{||}$ is the frictional force on the surface. For granular media the forces are found to be,
\begin{eqnarray}
    F &=& \int  2 k \rho g r |z| [f_\perp(\hat{v}.\hat{n})\hat{n} + (\hat{v}.\hat{t})\hat{t}]ds  \\
    f_\perp(\hat{v}.\hat{n}) &=& \left( 1 + \frac{C}{\sqrt{\tan^2\gamma_0 + (\hat{v}.\hat{n})^2} } \right)\hat{v}.\hat{n}
\end{eqnarray}
where $k\rho g |z|$ is the pressure at depth $|z|$ due to weight of material, with effective density $\rho$ and $k$ is a constant based on material (k=2.5 for glass particles with diameter 0.3mm). The least effort path is computed by minimizing total work done along the curve, from a given start pose to end pose obtained from demonstration. This is computed using the KKT conditions as follows,
\begin{eqnarray}
    w(g(x)) &=& \int F(x,\dot{x}) dx\\
    \min_{g(x)}w(x)&~&\text{subject to~~} h(x)\\
    \text{Solve} -\nabla w(x) &=& \sum^n_{j=1} \lambda_j h_j(x)
\end{eqnarray}
where $g(x)$ is the 6-DoF pose trajectory of the robot, $h(x)$ is the boundary conditions i.e. $h_1(x_1) = h_2(x_2) = 0$ where $x_1$ and $x_2$ are start and end poses of the trajectory in the media. Solving this gives the least effort path around which the action space for RL is defined. 
\subsection{Self-evaluation and Reinforcement learning} 
This paper involves executing a force based task, scooping. To help perform a task that uses force models to achieve the task, we use multiple sensing modalities as seen in Fig. \ref{fig:rob}. Vision sensors are used to detect the environment and help with the inference of the environment. Many tasks dealing with positional inference are easier to accomplish using vision as it depends on various optically salient object and state changes. However, in cases where a task requires other forms of sensor modalities to help make sense of the environment, vision will not suffice. Tasks involving contact or force models need additional force feedback which cannot be inferred from vision sensing. Thus we perform self-evaluation for any such skills where we need to tune parameters that cannot be inferred through vision alone.

Self-evaluation is done by means of Reinforcement Learning. In essence, the robot agent performs an action from a set of actions as part of its exploration to achieve a desired state change. Each action would generate a sensor-based feedback which would help model the reward. The goal of the  reinforcement learning algorithm is finding out the policy (set of actions to take) that maximizes the agents rewards. This potential reward is a weighted sum of the expected values of the rewards of all future steps starting from the current state. The learning is performed over a finite set of iterations. The states, actions and rewards enlisted below are the input to the learning algorithm.

\subsubsection{State  Space}
The state space within our RL framework is dependent on how we define our environment and the elements within that environment. The states are defined depending on the reward outcomes at each state. The goal state is already known to us to help tune the skill. In this paper, we deal with binary state space where we are either in the desired state which would provide us with a maximized reward that depends on the feedback obtained for the desired state, or alternatively, we are in a state which is not the desired state leading to a reward value that need not be zero but is definitely lesser than that of the reward obtained when present in the desired state.  

\subsubsection{Action Space}
Through the inference from a demonstration, we obtain a baseline trajectory for any skill that was demonstrated. The self evaluation RL framework tunes features in this trajectory and these features in their ranges represent the action space. The baseline trajectory is tuned to obtain the desired trajectory for the desired skill. This tuning (force tuning in our experiments) is performed due to the fact that vision cannot help estimate all the required information from the trajectory especially in tasks where other sensing models (apart from vision) are involved. The baseline trajectory is sampled into multiple points and on each point an action can be performed. In our experiment, each of these points can be tuned to an orientation to achieve the least resistive force along that point in the trajectory while doing the scooping action.  

\subsubsection{Rewards}
The reward for the reinforcement learner is a function of the force feedback obtained  from the environment after performing the desired action. This is different in comparison to using binary rewards where we can only hope to get a reward in the goal state. Since our defined state space is binary, we can model the reward function to provide a reward for cases where we are not in the desired goal state. The reward for the reinforcement learner is  a high positive reward if desired goal state $s*$ is reached, or, a lower positive  reward if it remains in the same state that is not the desired goal state. Let system transition from current state $s$ to new state $s'$ on performing action $a$. Then the reward is given by,
$$ R_a(s,s') = c_1*\delta_{s',s*} - c_2(1-\delta_{s',s*}) $$
where $\delta_{s',s*}$ is the Kronecker delta function resulting in $1$ when $s*$ and $s'$ are same and 0 otherwise and $c_2 > c_1>0$

\subsubsection{Q Learning}
Q learning \cite{Watkins1992} is a form of temporal difference learning. The motive of Q learning is to learn a policy that maximizes rewards over time by performing the best action at a given state. The algorithm uses $Q$ values which are defined for every state action pair. Each $q$ value is the maximum expected future reward for every state action pair. Q Learning has no defined policy initially and it works on improving the policy or achieving the desired policy over time by performing a set of state action  trials.  The best action at any state is the largest $q$ value corresponding to an action for that state in the Q table.\\
\begin{eqnarray}
    Q_{t+1}(s_t,a_t) &=& Q_{t}+\alpha[ r_{t+1}(s_t,a_t) + \gamma~ \max_a Q_t(s_{t+1},a) \nonumber \\&-& Q_{t}(s_t,a_t)]
\end{eqnarray}
In our application, we use the Q learning algorithm to perform self-evaluation and help tune our trajectories to optimize the skill in modalities that cannot be inferred through vision. Similar to the methods in \cite{balakuntala2019self} we have a new Q table that would be used for objects of certain classes which would be prior information obtained through our object database. Thus, we have multiple learnt Q tables depending on our interactions with the environment. 

The Q learning algorithm has two hyperparameters we deal with. Learning rate determines how much weight we give to the current $q$ value versus the older $q$ value. A large learning rate would provide the new $q$ value obtained with a higher weight while a lower learning rate would result in a  very slow learner. The second hyperparameter of importance is the discount rate. Discount factor determines how to give higher weight to near rewards received than rewards received further in the future. The reason for using discount factor is to prevent the total reward from going to infinity.

\subsection{Coaching} 
Once the skill is learnt by means of inference and self-evaluation, a task expert or coach would guide the robot to tune the skill to achieve a  user specific task goal that will be provided by the human in loop. Coaching is a means to only tune our already inferred skill policy to perform much better; it doesn'’t involve learning the whole skill policy again from demonstration or from inference.
\begin{figure}[t]
    \centering
        \centering
        \includegraphics[width=0.45\textwidth]{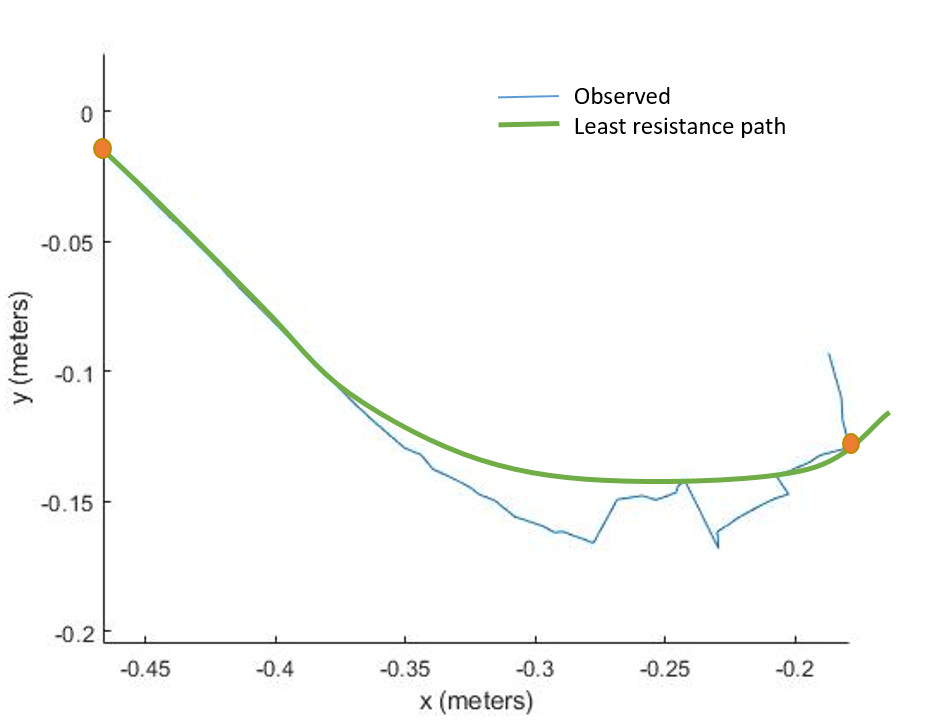}
        \caption{The least resistance path estimated using resistive force theory}
        \label{fig:lrpath}
\end{figure}
In the reinforcement learning framework used for coaching, we use a multi armed bandit approach, where the task expert is prompted by the RL agent to provide a specific task goal and  the required action space as the coaching input. These inputs are used to tune the learnt skill even further to achieve these user specific requests. The coaching feedback is a sense of direction for the robot agent guiding it to move along a path provided by the task expert. Since the robot only gets a vague idea of what is to be done from the coaching input, it uses the  Reinforcement Learning framework to help tune the policy to accomplish the desired user specified task goal. The specific task goal is a numeric input which would be treated as the desired state to be achieved and the rewards to the reinforcement learner would be modeled as a function of the feedback the robot obtains when attempting to tune its skill to achieve the desired state similar to the reinforcement learner used in the self-evaluation scenario.

Extrapolating from the example provided in section I, let us consider the example of teaching a robot to throw the basketball in the hoop.  We learn the skill to throw a basketball through inference from demonstration and self-evaluation. The task expert or the coach would then help us tune that skill(skill being throwing a ball ) to throw the ball to a specific location perfectly. The specific location will be the user specific task goal provided by the task expert and coached input will be the action space which in the case of throwing a ball would be the angle of the throw to help achieve various specific distances. The coached input would just mention if the angle should be higher or lower to achieve the user specific request. This input will lead us to define an action space that can be used to learn the policy to achieve the user specific request. In section IV we explain this with the help of the scooping  task chosen in this paper to provide more light on how a coaching input is provided in our experiments.

\section{Experiment}
To evaluate the extended one-shot learning approach, we use scooping granular media (steel ball bearings) as the task. The task involves scooping ball bearings from one container and unscooping into a different container. The policy is learnt as composition of sensorimotor skills some of which are tuned to achieved observed performance. The learnt policy is extended through coaching to scoop and unscoop desired amount of media. Scooping was chosen because it is a very common task in homes, kitchens, restaurants, and industries like in glove box operations. We consider the action of scooping to be the task learnt through the inference from demonstration and the coaching section would coach the robot on how to unscoop specific desired quantities provided by the coach.
\begin{figure}[t]
    \centering
        \includegraphics[width=0.45\textwidth]{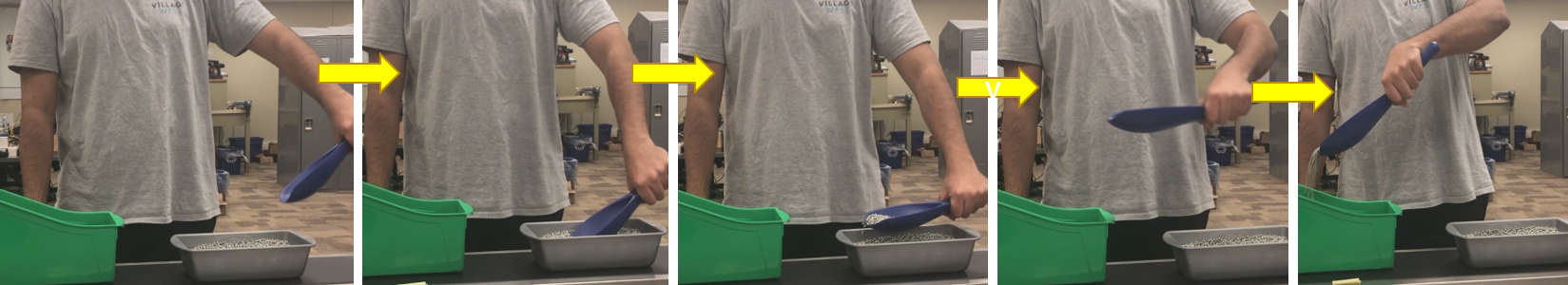}
    \caption{Progression of scooping demonstration}
    \label{fig:demo}
\end{figure}
\begin{figure}[t]
    \centering
        \includegraphics[width=0.45\textwidth]{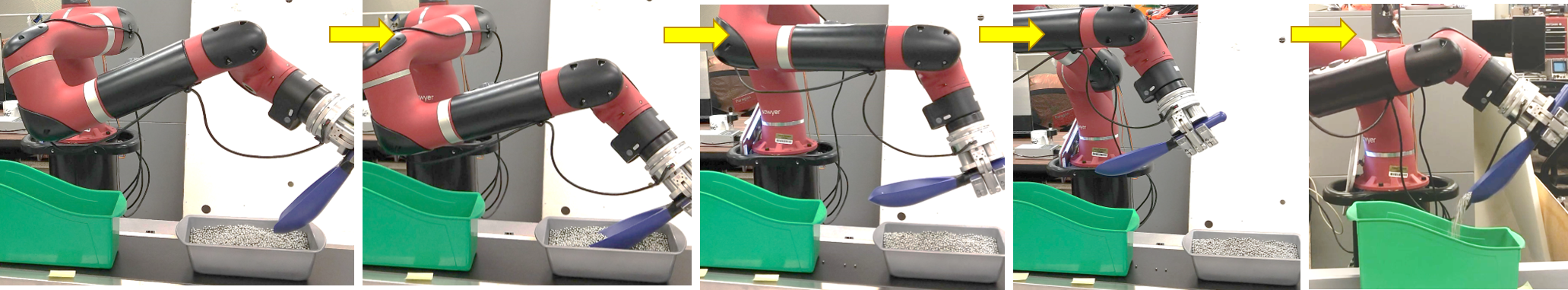}
    \caption{Execution of complete task}
    \label{fig:rob_exec}
\end{figure}
\subsection{Demonstration}
A single demonstration of the task is performed by the task expert. Importance is given to the task of scooping during inference which will then be tuned using self evaluation. The coach can choose to modify any of the skills, but here we use coaching to transfer desired amount of media. Hence the unscooping skill is modified using an RL framework through coaching The example of the demonstration used is shown in Fig. \ref{fig:demo}

\subsection{Inference} In the case of scooping task the policy learnt after identifying keypoints, segmenting, and classifying is,
\begin{eqnarray*}
 \Pi &=& \{(C_1, s*_1, \Theta_1), (C_2, s*_2, \Theta_2), (C_3, s*_3, \Theta_3),\\ &&(C_4, s*_4, \Theta_4), (C_5, s*_5, \Theta_5)\},
\end{eqnarray*} where skill $C_1$ is \textit{approach} container as reference, skill $C_2$ is \textit{scoop}, $C_3$ is \textit{lift}, $C_4$ is \textit{move to goal}, and $C_5$ is \textit{unscoop}. The states of the object, in this case, include position and orientation of tool w.r.t container and goal. The containers and the orientation of scoop is detected through vision and embedded encoders. The amount of material in hand is estimated using the 6-DoF Force/Torque sensor in the Barrett hand. 

\subsection{Execution}
For both the scoop and unscoop skills force sensor readings are used as feedback for the skill execution as described in the a priori skills section. The execution of the learnt policy is done based on the control policy described in section \ref{skills}. The control policy for approach is very straight forward, it reduces to a PD control based on visual error. The controller for $scoop$ uses a hybrid force controller and all other skills are kinematic. The $unscoop$ skill uses position controller but with force as the feedback to ensure the material was successfully unscooped. The execution of the learnt policy by the robot is shown in Fig. \ref{fig:rob_exec}.
For tuning the skill we use Q Learning bootstrapped with baseline least effort path which is shown in Fig. \ref{fig:lrpath}. 
\begin{figure}[t]
    \centering
    \includegraphics[width=0.46\textwidth]{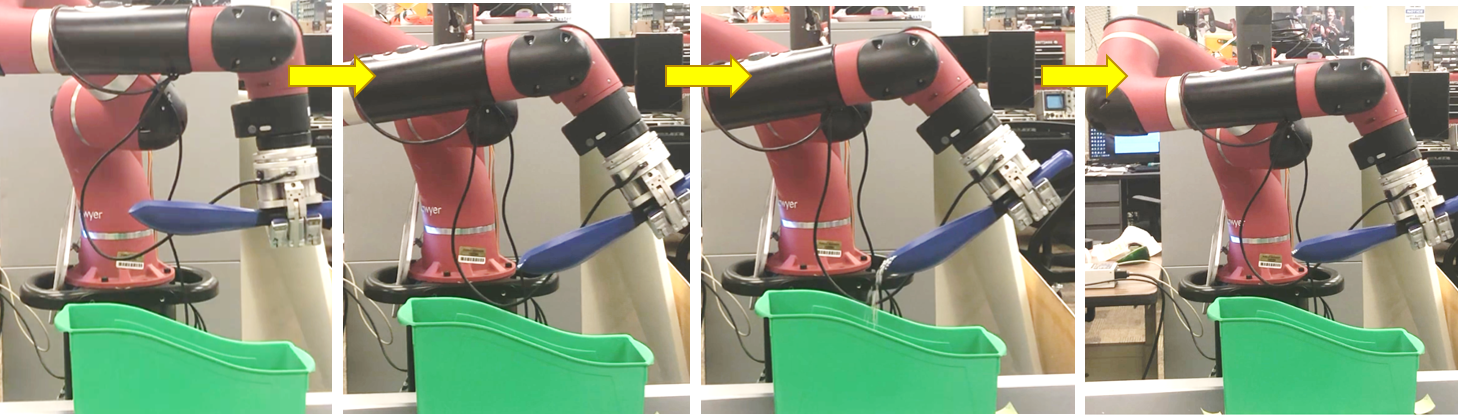}
    \caption{Unscooping after coaching, the robot orients appropriately to transfer desired amount and orients back}
    \label{fig:unscoop}
\end{figure}
\begin{figure}[t]
    \centering
    \includegraphics[width=0.46\textwidth]{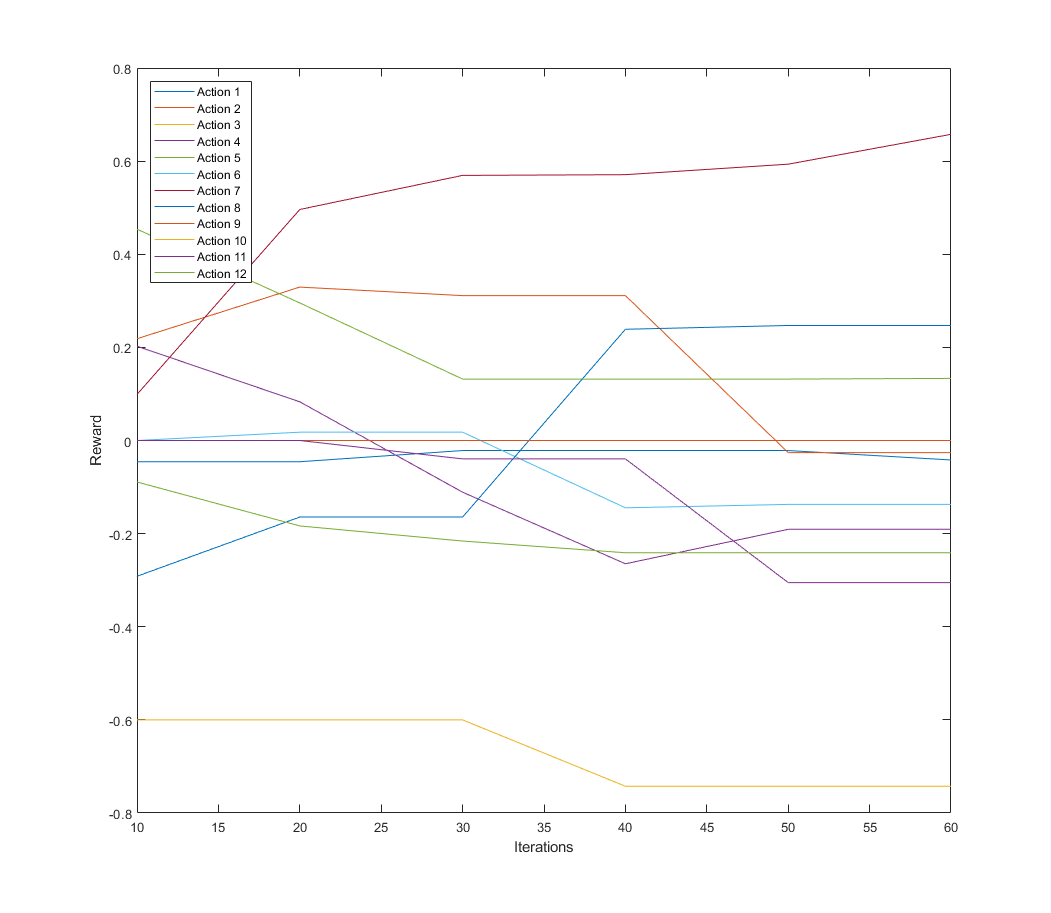}
    \caption{Expected rewards, for specified 100g material transfer}
    \label{fig:rew}
\end{figure}
\subsection{Self Evaluation}
The state space is essentially binary with two states in our specific experiment of scooping. The transition from the current state to the next state occurs when the robot performs the scooping task by experiencing the least effort which is the reward. The action space is not the trajectory inferred from the demonstration but rather discrete points along the trajectory that can orient themselves within a bounded action space. The goal of the self evaluation is to optimize the orientations across the trajectory such that the minimal effort is obtained when performing the tasks of scooping. Rewards is chosen to be a function of $w(g(x))$, which is the effort. 

\subsection{Coaching}
Once the robot performs self evaluation and learns to perform the task of scooping, it prompts the task expert for the task specific input to achieve the task specific goal which in our case is to unscoop a desired quantity. The task expert would enter a desired weight quantity which is a numeric value the robot would be looking to unscoop. The robot agent would then prompt the task expert for the coaching input. This input can be one of seven possible values ${x,y,z,roll,pitch,yaw}$ each associated with a particular degree of motion for the robot. In our case, the coach input specified for unscooping by the task expert is the $pitch$ angle. The coach input also specifies what direction the action space is to be defined in depending on the degree of motion chosen $(pitch)$. In our case the task expert provides the information 'down'. Once the coach input $(pitch,down)$ is provided we define our action space across that input space. The starting position and orientation of the hand in our action space is obtained from the demonstration when the user moves to a start location after scooping. Once the action space and unscoop quantity (task specific goal) has been provided, the robot agent uses reinforcement learning to tune its orientation to achieve the desired unscooping quantity. Fig. \ref{fig:unscoop} shows the robot performing the unscooping task. Using the task experts inputs provided we were able to coach the robot to unscoop desired quantities as required. The learning algorithm used an epsilon approach to handle exploration vs exploitation. Epsilon was modeled such that initially we have a higher weight towards exploration and, as the epochs increase, we provide higher weight to exploitation.  Convergence was obtained within 30 iterations as seen in Fig. \ref{fig:rew}. The robot managed to unscoop the desired quantity specified by the user through the proposed human in loop reinforcement learning framework by means of coaching. 

\section{Conclusions}

In this paper, we presented a coaching framework to extend policies learnt through  one-shot learning from demonstration using our Super Baxter robot. Reinforcement learners are used to perform self-evaluation which is used to tune a set of skills. The goal is to perform such skills as observed and provide coaching through feedback from task experts. Such feedback is used to extend the learnt policy to achieve desired task goals. Coaching provides the task expert with the ability to coach the robot to extend the policy  learnt. The policy is extended so as to  be  adaptable to specified task  goals, in turn, enabling generalization of the learnt model. The proposed framework was evaluated using the task of scooping and unscooping, where the desired task goal was to transfer a specific amount of granular material. The framework uses force models to bootstrap self-evaluation RL to reduce the search space hence, enabling faster convergence to observed policy. The self-evaluation was able to learn the observed policy from one-demonstration and coaching. The transferred specific amount in the scooping task was seen to converge within 30 trails, signifying the success of the proposed method. The presented coaching methods implemented in this paper can be used on any task to extend the task capabilities or to correct suboptimal policies by guided feedback from the task experts.

\bibliographystyle{IEEEtran}
\bibliography{refs_coaching}
\end{document}